\documentclass{./styles/svproc}
\usepackage{url}
\usepackage{graphicx}
\usepackage{caption,subcaption}
\usepackage{multirow}

\begin{document}
\mainmatter              
\title{Comparative Analysis of Imbalanced Malware Byteplot Image Classification using Transfer Learning}
\titlerunning{Malware Classification}  
%
\author{Jayasudha M, Ayesha Shaik, Gaurav Pendharkar, Soham Kumar,
Muhesh Kumar B, Sudharshanan Balaji}
\authorrunning{Jayasudha M et al.} 
%
\tocauthor{Jayasudha M, Ayesha Shaik, Gaurav Pendharkar, Soham Kumar,
Muhesh Kumar B, Sudharshanan Balaji}
\institute{Vellore Institute of Technology,  Chennai-600127, Tamil Nadu, India,\\
\email{\{jayasudha.m,ayesha.sk\}@vit.ac.in},\\ \{gauravsandeep.p2020, soham.kumar2020, muheshkumar.b2020, sudharshanan.pb2020\}@vitstudent.ac.in\\}

\maketitle              

\begin{abstract}
Cybersecurity is a major concern due to the increasing reliance on technology and interconnected systems. Malware detectors help mitigate cyber-attacks by comparing malware signatures. Machine learning can improve these detectors by automating feature extraction, identifying patterns, and enhancing dynamic analysis. In this paper, the performance of six multiclass classification models is compared on the Malimg dataset, Blended dataset, and Malevis dataset to gain insights into the effect of class imbalance on model performance and convergence. It is observed that the more the class imbalance less the number of epochs required for convergence and a high variance across the performance of different models. Moreover, it is also observed that for malware detectors ResNet50, EfficientNetB0, and DenseNet169 can handle imbalanced and balanced data well. A maximum precision of 97\% is obtained for the imbalanced dataset, a maximum precision of 95\% is obtained on the intermediate imbalance dataset, and a maximum precision of 95\% is obtained for the perfectly balanced dataset.
\end{abstract}
\section{Introduction}
Cyber threats are still a big issue for people, businesses, and governments all around the world. The growing reliance on technology and networked systems has increased the sophistication and prevalence of cyberattacks, posing a serious danger to data security and privacy.\cite{bib1} Phishing, ransomware, and distributed denial-of-service (DDoS) attacks are among the most typical forms of cyber attacks. These attacks can cause significant financial and reputational damage, disrupt essential services, and even pose a threat to national security. In response, there has been a growing emphasis on cybersecurity measures, including the adoption of advanced encryption technologies and the implementation of comprehensive cyberdefense strategies.

A malware detector is a software tool designed to identify and remove malicious software, also known as malware, from a computer or network. The detector typically works by scanning files and system components for suspicious behavior or known patterns of malicious code. Malware Analysis typically involves two techniques, static analysis, and dynamic analysis. Static analysis is a method used to examine the behavior and structure of a malware sample without executing it. This type of analysis involves analyzing the binary code of the malware and identifying its various components, such as function calls, system calls, libraries imported, and metadata like file size, timestamps, and digital signatures.\cite{bib2} Dynamic malware analysis involves running the malware in a controlled environment to observe its behavior, which helps to identify the malicious actions that it performs on a system. \cite{bib2}

Traditional malware detection techniques rely on signature-based detection, which can be limited in its ability to detect new and emerging threats. Machine Learning can help in detecting previously unknown malware by identifying subtle patterns in the malware behavior. It can automate the process of feature extraction by converting malware files into byteplot representations. Furthermore, it can improve dynamic analysis by identifying suspicious behavior patterns in real-time and flagging potentially malicious activities and software. Overall, machine learning can make malware detection more efficient, robust, and sophisticated.

The malware byteplot image datasets used for the proposed work are the Malimg dataset\cite{bib3} and the Malevis dataset\cite{bib4}. A hybrid dataset is created by blending the two open-source datasets. Moreover, a comparative analysis is carried out on the three datasets to acquire insights into the effect of class imbalance on malware byteplot image classification. The state-of-the-art CNNs are used to achieve the multiclass classification of malware and compare their performance. 

The comparative analysis will foster the development of machine learning-based malware detectors by helping to choose the right model based on the ability of the model to handle class imbalance.
\section{Related Work}
The multiclass classification of malware byteplot images has been tried in literature by using various data augmentation techniques, sequential modeling, and convolutional neural networks. Agarap, A. F. et al., 2017 discuss an SVM-based deep learning model to classify the byteplot images in the Malimg dataset with various feature extractors like MLP, CNNs, and GRUs. They achieve a predictive accuracy of 84.92\% with the Malimg dataset and GRU-SVM model. The usage of sequential models to process the byteplot images of varied sizes is commendable but the accuracy is comparatively decent.\cite{bib5} Kalash, M. et al., 2018 design a CNN-based framework that is proposed to render better performance than the traditional approaches of shallow learning for malware classification using Byteplot images. They test the model on the Malimg dataset and Microsoft dataset resulting in an accuracy score of 98.52\% and 99.97\% accuracy respectively. The accuracy of the customized framework is commendable but on the other hand, only accuracy is used as the primary metric on an imbalanced dataset like the Malimg dataset.\cite{bib6}

Lo, W. W. et al., 2019 discuss an Xception model that performs better than existing models like VGG16 and other traditional models like KNN and SVM. XceptionNet obtains the highest validation accuracy against the other models VGG16, KNN, and SVM. The high accuracy is commendable but the usage of accuracy as the primary metric can be misleading on the actual nature of the model.\cite{bib7} Singh, A. et al., 2019 prepare a malware dataset using data collection, and deep neural networks are designed to classify the images across 22 families. An accuracy of 98.98\% and 99.40\% using deep CNN and ResNet-50 respectively is achieved. The high accuracy is commendable.\cite{bib8}

J. H. Go et al., 2020 experiment ResNeXt model for the classification of malware byteplot images. They achieve an accuracy of 98.32\% and 98.86\% on the  Malimg dataset and Malimg dataset after image enhancement. The enhancement of image quality and the resulting high accuracy is commendable but accuracy cannot be a sufficient metric to evaluate the model quality for an imbalanced dataset like the Malimg dataset.\cite{bib9} Ghouti et al., 2020 discuss an approach of extracting image features after a Principal Component Analysis and then using an SVM to perform the classification. They use the Malimg, Ember, and BIG 2015 malware datasets to reach accuracy values of 99.8\%, 91.1\%, and 99.7\%, respectively. Evaluation of the model on different datasets gives to a good understanding of the model quality but dimensionality reduction can lead to the loss of information.\cite{bib10} Mitsuhashi, R. et al., 2020 discuss an approach to solve the data imbalance using the undersampling technique and fine-tuning VGG19 on the Malimg dataset. They obtained an accuracy of 99.72\%. The high accuracies and data augmentation is commendable but the usage of accuracy as the primary metric can be misleading for an imbalanced dataset like the Malimg dataset.\cite{bib11} Danish Vasan et al., 2020 experiment with transfer learning using the Malimg dataset and IoT-android mobile dataset. The performance of this model is compared with existing pre-trained CNNs. The Malimg malware dataset shows accuracy of 98.82\%, and the IoT-android mobile dataset shows accuracy of about 97.35\%. The high accuracies are commendable but the usage of accuracy as the primary metric can be misleading for an imbalanced dataset like the Malimg dataset.\cite{bib12}

Aslan, Ö. et al., 2021 discuss a hybrid model integrating the performance of two pre-trained models namely AlexNet and ResNet152 in an optimal manner. The model is tested on Malimg, Microsoft BIG 2015, and Malevis datasets. For the Malimg dataset, it gives 97.78\% accuracy. The higher accuracy and usage of a hybrid model are commendable but the usage of accuracy as the primary metric can be misleading for an imbalanced dataset like the Malimg dataset.\cite{bib13} Asam, M. et al., 2021 discuss an approach to the extraction of features from multiple CNNs and fusing their results. Finally, using an SVM to discriminate between them. The architecture achieves an accuracy of 98.61\%, an F-score of 0.96, a precision of 0.96, and a recall of 0.96. The performance of the model is good and its evaluation using different classification metrics is commendable.\cite{bib14} Awan, M.J.  et al., 2021 discuss a spatial attention and convolutional neural network approach for the multiclass classification of malware. They achieve a precision of 97.42\%, a recall of 97.95\%,  a specificity of 97.33\%, and an F1 score of 97.32\%. The performance of the model is good based on the reported classification metrics.\cite{bib15}

Mallik, A. et al., 2022 describe an approach to resolve data imbalance using data augmentation and the augmented dataset is classified by using 2 LSTM layers and 1 VGG Net. The overall results from each are integrated and combined. Treating the malware file bits as a bidirectional dependency is commendable.\cite{bib16} AlGarni, M. D.  et al., 2022 have compared the performance of EfficientNetB3 on Imagenet and Malimg datasets. They obtain an accuracy of 99.93\% on the Malimg dataset. The comparison of the performance of the model on two datasets is commendable but accuracy cannot be a sufficient metric to evaluate the model quality for an imbalanced dataset like the Malimg dataset.\cite{bib17} Adem Tekerek et al., 2022 resolve the classification by data augmentation using CycleGAN and use different CNNs for classification. They achieve an accuracy of 99.86\% for the BIG2015 dataset and 99.60\% for the Dumpware10 dataset. The high accuracy is commendable.\cite{bib18}
\section{Proposed Architecture}
\begin{figure}[htbp]
\centering
\includegraphics[width=1\textwidth]{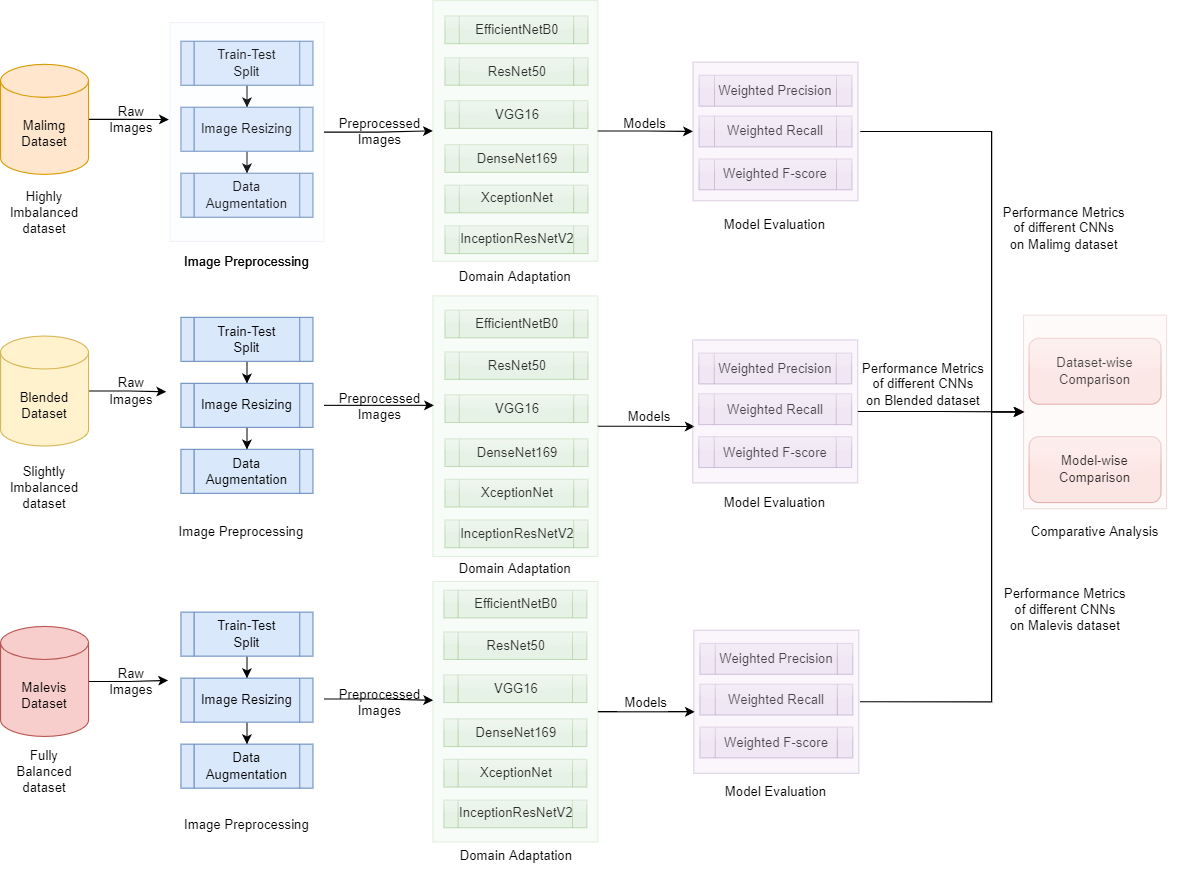}
\caption{Architecture Diagram}
\label{arch-diag}
\end{figure}

Figure \ref{arch-diag} shows the architecture diagram for the flow of data for the comparison of the imbalanced image classification of three different malware datasets. Two malware image datasets are available namely the Malimg dataset and the Malevis dataset. Both datasets are blended into a single dataset of intermediate imbalance. All the images from the respective datasets are subject to an initial Image Preprocessing comprising Image Resizing and Augmentation. At the end of this stage, there are three splits available for each dataset: train, validation, and test. Following this, a set of six models are experimented on each of the datasets and evaluated based on Weighted Precision, Weighted Recall, and Weighted F-score. The performance metrics for each of the models are taken into account for comparative analysis of the variation of model performance based on malware class imbalance. The models were trained using GPU P100. In the forthcoming sections, each of the steps is discussed in detail.
\section{Proposed Methodology}
\subsection{Data Blending}
\begin{figure}
\centering
    \begin{subfigure}[b]{3.15in}
\includegraphics[width=\textwidth]{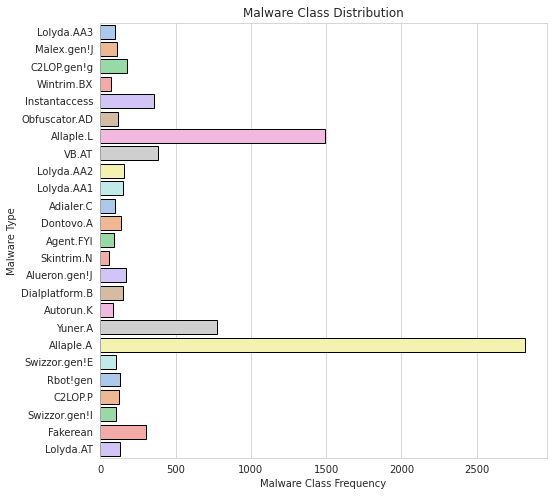}
    \caption{Malimg dataset}
    \end{subfigure}\hfil
    \begin{subfigure}[b]{3.15in}
\includegraphics[width=\textwidth]{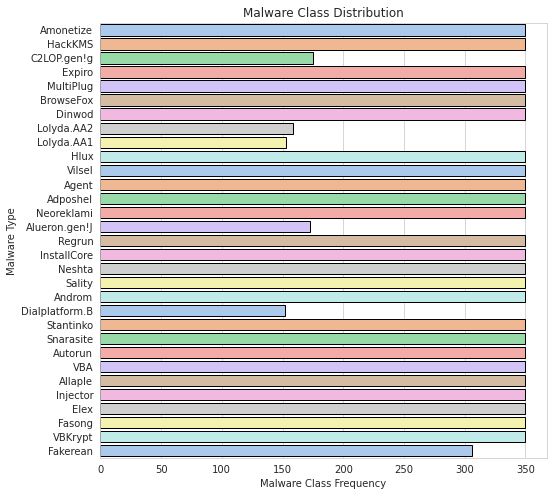} 
   \caption{Blended dataset}
    \end{subfigure}

    \begin{subfigure}[b]{3.15in}
\includegraphics[width=\textwidth]{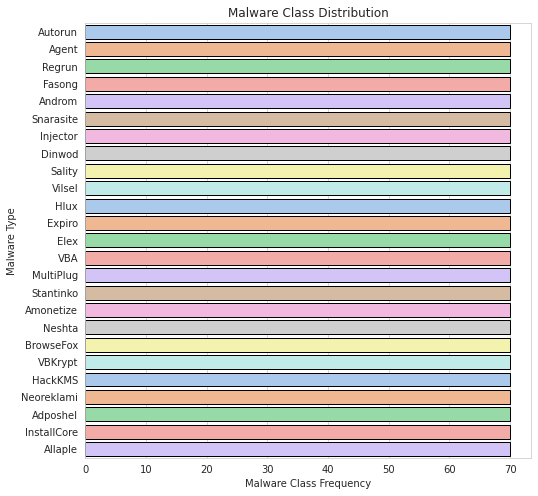}
   \caption{Malevis dataset}
    \end{subfigure}\hfil
\caption{Class Distribution of the three datasets}
\label{dataset-class-distributions}
\end{figure}

The act of merging data from many sources, sometimes with different formats or structures, to produce a single dataset that can be utilised for analysis is known as data blending. A blended dataset is created by blending 5 major classes from the Malimg dataset into the 25 malware classes of the Malevis dataset. Finally, three datasets are obtained namely the Malimg dataset as a fully imbalanced dataset, the Blended dataset as a dataset of intermediate imbalance, and the Malevis dataset as a perfectly balanced dataset. The class distribution of the datasets is shown in Figure \ref{dataset-class-distributions} using the bar charts.

\subsection{Image Preprocessing}
\begin{figure}[htbp]
\centering
\includegraphics[width=1.1\textwidth]{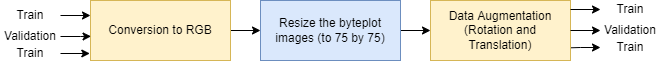}
\caption{Image Preprocessing}
\label{Image Preprocessing}
\end{figure}
\begin{figure}
\centering
    \begin{subfigure}[b]{2.5in}
\includegraphics[width=\linewidth]{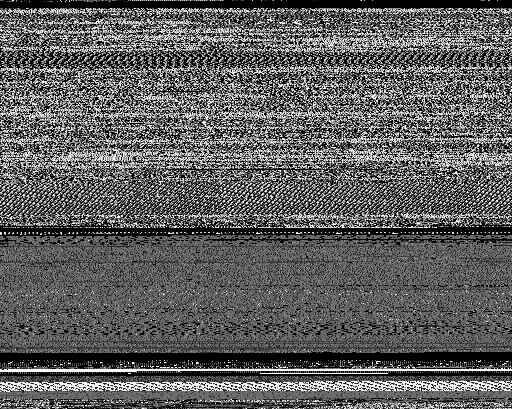}
    \caption{Grayscale Byteplot Image}
    \end{subfigure}\hfil
    \begin{subfigure}[b]{2.1in}
\includegraphics[width=\linewidth]{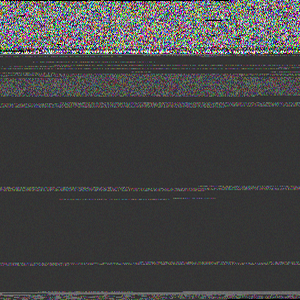} 
   \caption{RGB Byteplot Image}
    \end{subfigure}
\caption{Types of byteplot images}
\label{byteplot-images}
\end{figure}

Image Preprocessing is a preliminary step to normalize and augment the image data before feeding it into a neural network for training. Among the datasets, the Malimg dataset comprises grayscale images and the Malevis dataset comprises RGB images as shown in Figure \ref{byteplot-images}. Consequently, the Blended dataset consists of both grayscale and RGB images. Moreover, the Malimg dataset has all the images of different sizes which need to be converted to a single image size. After the data is split into a train, test, and validation, all the images are converted RGB and resized to 75 by 75. Following this, the images are augmented by rotating and translating the images which make the model rotation and translation invariant. The whole process for image preprocessing is shown in Figure \ref{Image Preprocessing}.

\subsection{Domain Adaptation}
Domain adaptation in transfer learning refers to the process of using knowledge gained from one domain to improvise the model performance in a different but related domain. It involves transferring knowledge from a source domain to a target domain, where the data distributions may be different. The data distribution of byteplot images is learned by the model after making the last few feature extraction layers trainable. In addition to it, dropout regularization layers are added to the architecture to mitigate the chance of overfitting for the fully balanced dataset. Moreover, the models are trained with Early Stopping by monitoring the validation loss of each epoch. These additional components to architecture make it less prone to overfitting.

For each of the datasets, six state of art convolutional neural networks were experimented particularly XceptionNet\cite{bib19}, EfficientNetB0\cite{bib20}, ResNet50\cite{bib21}, \\DenseNet169\cite{bib22}, VGG16\cite{bib23}, and InceptionResNetV2\cite{bib24}.
\subsection{Evaluation Metrics}
A machine learning evaluation metric is a numerical measure that is used to evaluate the effectiveness of a machine learning model. It is used to evaluate how well the model's predictions match the actual outcomes or labels of the data used to train and test the model. For an imbalanced multiclass classification problem, the common accuracy metric is not appropriate since does not take into consideration the support images available across each of the classes. Hence, the most appropriate metrics for the evaluation of each of the six models are weighted precision, weighted recall, and weighted F1-score as depicted in the equations \ref{precision}, \ref{recall}, and \ref{f1-score} respectively. In the test data, every class has a specific number of images for evaluation \texttt{$w_{i}$} and the precision \texttt{$p_{i}$}, recall \texttt{$r_{i}$}, and F1-score \texttt{$f1_{i}$} on comparison with the ground truth.

\begin{equation}
Weighted \hspace{2pt} Precision \hspace{2pt} = \hspace{2pt} \frac{\sum_{i \epsilon C }^{} w_{i}*p_{i}}{\sum_{i \epsilon C }^{} w_{i}}
\label{precision}
\end{equation}

\begin{equation}
Weighted \hspace{2pt} Recall \hspace{2pt} = \hspace{2pt} \frac{\sum_{i \epsilon C }^{} w_{i}*r_{i}}{\sum_{i \epsilon C }^{} w_{i}}
\label{recall}
\end{equation}

\begin{equation}
Weighted \hspace{2pt} F1-score \hspace{2pt} = \hspace{2pt} \frac{\sum_{i \epsilon C }^{} w_{i}*f1_{i}}{\sum_{i \epsilon C }^{} w_{i}}
\label{f1-score}
\end{equation}

The use of weighted metrics helps in standardizing the performance of models across datasets of different extents of imbalance. In the forthcoming section, the models and performance are compared using these metrics.
\section{Results and Discussion}
In most cases, Machine Learning based Malware detectors are the multiclass classifiers. In this paper, the focus is on the comparison of multiclass classification of malware byte plot images on three different datasets. Table \ref{eval-metrics} shows the evaluation metrics for six CNNs across three datasets. From the results, it is evident that the more balanced the dataset is less is the variance in the performance of models. In the Malimg dataset, there is a high variance across the evaluation metrics of the six models. The best performance is achieved by EfficientB0 with a precision of 97\%, recall of 96\%, and F-score of 96\%. In the case of the Blended dataset, the best performance is achieved by ResNet50 with precision, recall, and an F-score of 95\%. For Malevis Dataset, almost all models perform well because of the balance in its class distribution. However, XceptionNet, EfficientB0, and DenseNet169 are performing the best with precision, recall, and an F-score of 95\%.
\begin{table}
\begin{center}
\caption{Evaluation metrics for CNNs (in \%)}\label{eval-metrics}
\begin{tabular}{clllllllll}
\hline
\multirow{2}{*}{Model}           & \multicolumn{3}{c}{Malimg}                 & \multicolumn{3}{c}{Malevis}       & \multicolumn{3}{c}{Blended}          \\
                  & P      & R & F1 & P       & R & F1 & P      & R & F1  \\
\hline
XceptionNet       & 87             & 86     & 85      & 95              & 95     & 95      & 92              & 92     & 92       \\
EfficientNetB0    & 97             & 96     & 96      & 95              & 95     & 95      & 92              & 92     & 92       \\
ResNet50          & 95             & 95     & 95      & 93              & 93     & 93      & 95              & 95     & 95       \\
VGG16             & 79             & 80     & 79      & 93              & 92     & 92      & 92              & 92     & 92       \\
DenseNet169       & 95             & 96     & 95      & 95              & 95     & 95      & 94              & 94     & 94       \\
InceptionResNetV2 & 91             & 91     & 91      & 94              & 93     & 93      & 93              & 93     & 93 \\
\hline
\end{tabular}
\end{center}
\end{table}
\subsection{Comparison across models}
\begin{figure}
\centering
    \begin{subfigure}[b]{2.1in}
\includegraphics[width=\linewidth]{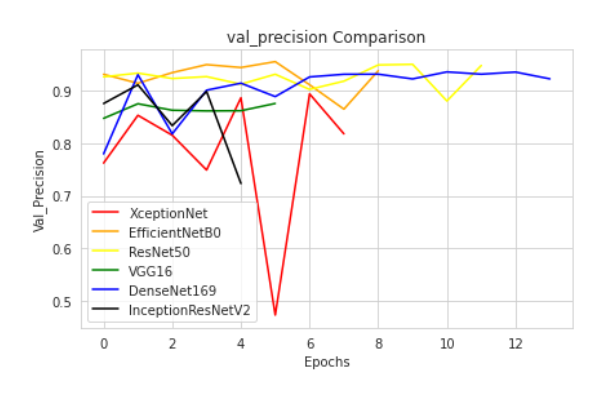}
    \caption{Malimg Dataset}
    \end{subfigure}\hfil
    \begin{subfigure}[b]{2.1in}
\includegraphics[width=\linewidth]{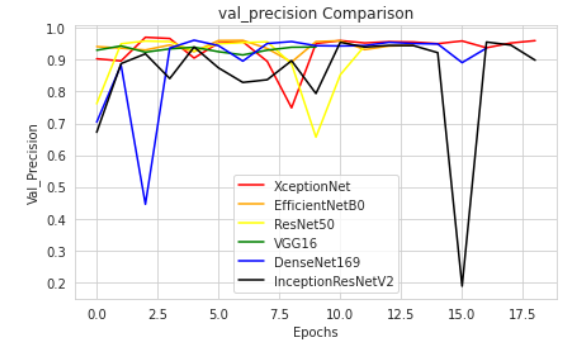} 
   \caption{Blended Dataset}
    \end{subfigure}
    \begin{subfigure}[b]{2.1in}
\includegraphics[width=\linewidth]{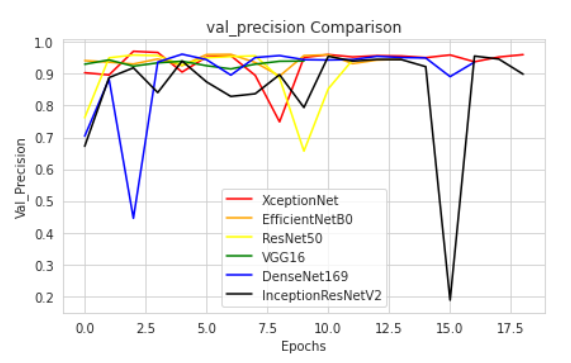} 
   \caption{Malevis Dataset}
    \end{subfigure}
\caption{Model-wise comparison of validation precision metric while training}
\label{model-wise-comparison}
\end{figure}

Precision is the most important metric for a malware detector because false positives turn out to be more expensive than a false negatives. Therefore, alterations of the validation precision are examined over the time of all the epochs as shown in Figure \ref{model-wise-comparison}. XceptionNet performs well for each epoch for the blended dataset and malevis dataset but is not able to learn well from imbalanced data. ResNet50, EfficientNetB0, and DenseNet169 both perform well with balanced as well as imbalanced data. VGG16 does not learn from imbalanced data but performs well for balanced data. InceptionResNetV2  has decent overall performance but its training history has a lot of spikes in validation loss and evaluation metrics making it unreliable. 

\subsection{Comparison across datasets}
\begin{figure}
\centering
    \begin{subfigure}[b]{2.1in}
\includegraphics[width=\linewidth]{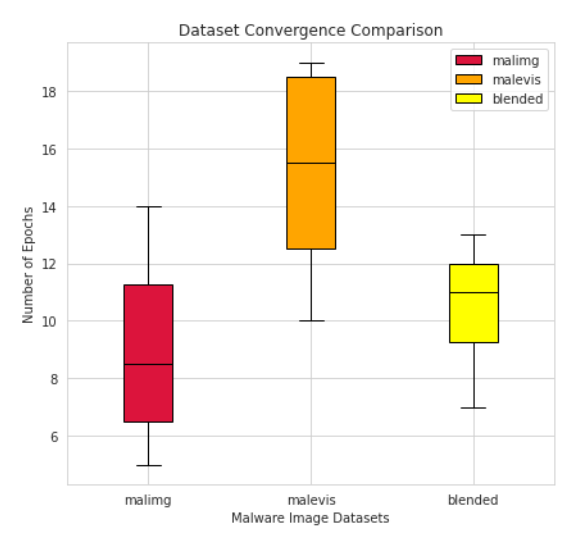}
    \caption{Convergence of models}
    \end{subfigure}\hfil
    \begin{subfigure}[b]{2.1in}
\includegraphics[width=\linewidth]{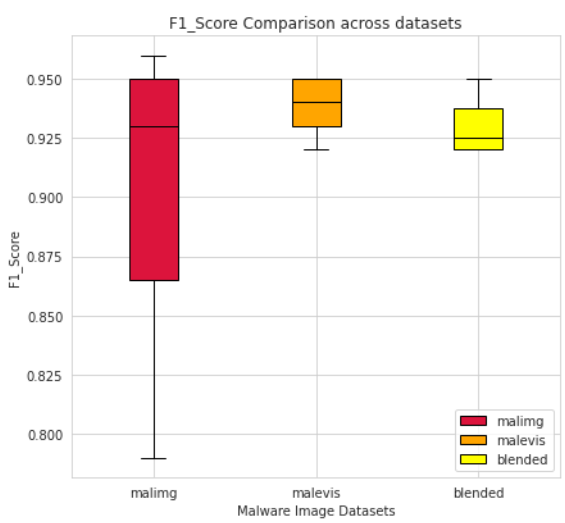} 
   \caption{F1-score distribution}
    \end{subfigure}\hfil
\caption{Dataset-wise comparison}
\label{dataset-wise-comparison}
\end{figure}

Previously the comparison was done by comparing the performance of different models on each of the datasets. Now, a comparison is carried out based on the datasets. The boxplot for model convergence is basically based on the distribution of the number of epochs required by each of the models and the distribution of the F1-score is also examined as shown in Figure \ref{dataset-wise-comparison}. From the convergence boxplot, it's evident that the Malimg dataset takes minimum epochs for convergence and the Malevis dataset takes the maximum epochs for convergence. The median epochs and median F1-score are represented by the horizontal line in the box which is perfectly in the center for the perfectly balanced and most deviated for the unbalanced dataset.
Moreover, there is a high variance in the performance of the models for the Malimg dataset compared to the other datasets. However, the evaluation metrics are one of the aspects but the size of the model and complexity must also be taken into account. Overall, it is evident that the imbalance in the class distribution directly affects the convergence of the model and its performance.

\subsection{Comparison with existing benchmarks}
From the literature survey, it is seen that most of the papers consider accuracy as the primary metric for an imbalanced malware dataset like the Malimg dataset. Accuracy is not the appropriate metric with reference to imbalanced classification resulting in model evaluation biased to the majority classes. Alternatively, weighted precision and weighted recall can serve as the primary metric. The F-score shall be used to condense the precision and recall into a single metric to foster model selection. In the literature, the precision obtained on the Malimg dataset is between 97\% and 98\% which is very close to the performance of EfficientNetB0 on that dataset. However, on Malevis dataset the precision ranges from 96\% to 98\% which is slightly higher than the maximum precision of 95\% obtained on this dataset.  

This paper contributes a blended dataset and the results for the same are a new finding. Moreover, a comparison of transfer learning on state-of-the-art CNNs serves as a head start to designing new malware detectors by reducing the experimentation required for model selection. Consequently, it saves the available hardware resources that can be utilized for more intensive tasks. 
\section{Conclusion}
A blended dataset is created by the data blending of the Malimg dataset and the Malevis dataset. The newly prepared dataset has an intermediate class imbalance compared to the two parent datasets. Finally, a maximum precision of 97\% is obtained for the imbalanced dataset, a maximum precision of 95\% is obtained for the intermediate imbalance dataset, and the perfectly balanced dataset. From the comparative analysis, it is observed that the more the class imbalance in the dataset more is the variance in the performance of different models and the number of epochs required for convergence. Moreover, it is also observed that for malware detectors ResNet50, EfficientNetB0, and DenseNet169 can handle imbalanced and balanced data well. On the other hand, VGG16 and XceptionNet were sensitive to class imbalance. This comparative analysis can help in choosing the models for experimentation while training any machine learning-based malware detectors.

\end{document}